\documentclass[a4, 10 pt, conference]{ieeeconf} 

\IEEEoverridecommandlockouts  

\usepackage{graphics} 
\usepackage{epsfig} 
\usepackage{mathptmx} 
\usepackage{times} 
\usepackage{amsmath} 
\usepackage{amssymb}  

\usepackage{booktabs} 
\usepackage[whole]{bxcjkjatype} 
\usepackage{hyperref} 

\usepackage{multirow}

\title{\LARGE \bf
Team Flow at DRC2022: Pipeline System for Travel Destination Recommendation Task in Spoken Dialogue
}

\author{Ryu Hirai$^{1*}$, Atsumoto Ohashi$^{1*}$, Ao Guo$^{1}$, Hideki Shiroma$^{1}$, Xulin Zhou$^{1}$, \\ Yukihiko Tone$^{1}$, Shinya Iizuka$^{2}$ and Ryuichiro Higashinaka$^{1}$ 
\thanks{$^{1}$ Graduate School of Informatics, Nagoya University, Japan
{\tt\small
hirai.ryu.k6@s.mail.nagoya-u.ac.jp
}
} %
\thanks{$^{2}$ School of Informatics, Nagoya University, Japan
}
\thanks{$^{*}$ Equal contribution.} %
}

\begin{document}

\maketitle
\thispagestyle{empty}
\pagestyle{empty}

\begin{abstract}
To improve the interactive capabilities of a dialogue system, e.g., to adapt to different customers, the Dialogue Robot Competition (DRC2022) was held. As one of the teams, we built a dialogue system with a pipeline structure containing four modules. The natural language understanding (NLU) and natural language generation (NLG) modules were GPT-2 based models, and the dialogue state tracking (DST) and policy modules were designed on the basis of hand-crafted rules. After the preliminary round of the competition, we found that the low variation in training examples for the NLU and failed recommendation due to the policy used were probably the main reasons for the limited performance of the system.
\end{abstract}

\section{INTRODUCTION}
With the popularization of human-machine dialogue, a dialogue system is expected to achieve objectives in various situations, e.g., respond to different customers appropriately in a customer service task. To improve the interactive capabilities of a dialogue system, Dialogue Robot Competition 2022 (DRC2022) \cite{drc2022} was held following the past competition \cite{drc2021}. Each team was required to develop a dialogue system embedded within a humanoid robot to handle the ``travel destination recommendation task.'' In the task, the robot plays the role of a counter-sales person to satisfy the customer by helping him/her choose one of two tourist attractions.

This paper reports the work of the team ``Flow'' in DRC2022. Our dialogue system was built with a pipeline composed of four modules: natural language understanding (NLU), dialogue state tracking (DST), policy, and natural language generation (NLG). By configuring the system in a pipeline fashion, (1) it is easy to tune the functionality of each module, and (2) in the future, we can expect to introduce a method, such as \cite{ohashi-2022-post}, to integrate all modules and optimize the dialogue performance of the entire system. We built the NLU and NLG modules by fine-tuning GPT-2 \cite{radford2019language}, a popular large-scale language model, with our crowdsourced data for the travel destination recommendation task. We further designed the DST and policy modules with hand-crafted rules.

After the preliminary round of the competition, we examined the evaluation results and dialogue histories. We found two main reasons for the limited performance: (1) the NLU was not able to track the customer dialogue acts properly due to the low variation in training examples for GPT-2, and (2) the rules of the policy resulted in a recommendation strategy that ignored customers' preferences.

\section{IMPLEMENTATION}
\begin{figure}[t]
\centering
\includegraphics[scale=0.27]{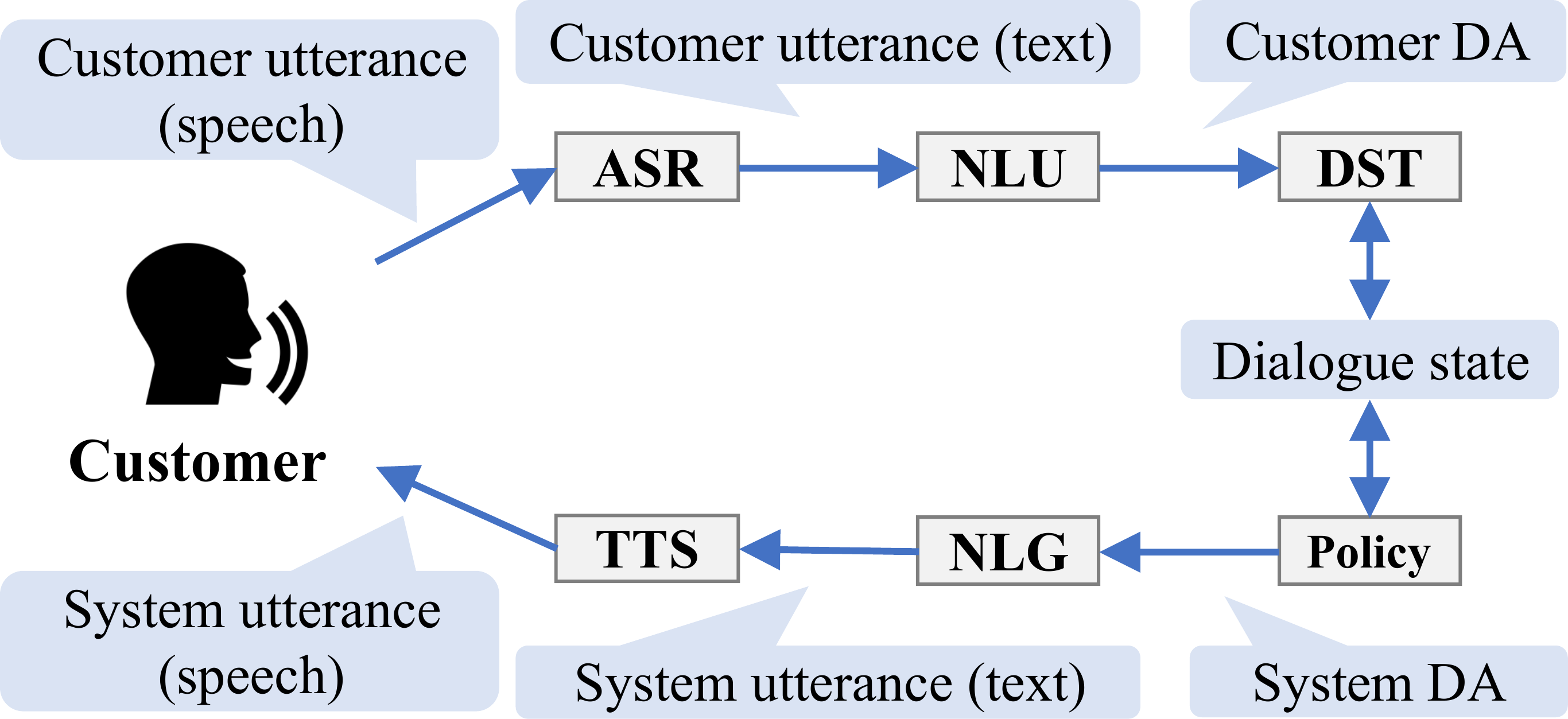}
\caption{Diagram of pipeline structure of our spoken dialogue system. At each turn, customer's speech recognition results obtained by automatic speech recognition (ASR) are processed by NLU, DST, policy, and NLG to generate system's response text, which is finally converted to speech by text-to-speech (TTS) to respond to customer.}
\label{fig:pipeline_architecture}
\end{figure}

Fig. \ref{fig:pipeline_architecture} shows the pipeline structure of the spoken dialogue system our team implemented. At each turn, the customer's speech input to the robot is converted into text by the automatic speech recognition (ASR) module, and the utterance text is input to the NLU module. The NLU predicts the customer's dialogue act (DA), which is a semantic representation of the customer's utterance. The DST module then updates the dialogue state on the basis of the customer's DA. The dialogue state consists of information such as the history of the DAs, the customer profile, and the \emph{belief state}, which is a set of customers' preferences toward travel. The policy module decides the next action to be taken by the system as the system's DA by using the dialogue state and information on tourist attractions from the database. The NLG module then converts the system's DA into a system utterance. Finally, the text-to-speech (TTS) module responds to the customer by converting the text response to speech. ASR and TTS were implemented using the Google Speech Recognition system and Amazon Polly API, respectively, which were prepared by the competition organizers. The robot's expression control and motion control were based on the expression and motion rules defined for each system dialogue act.

In the following sections, first, we describe the ontology of DA for the travel destination recommendation task, NLU, DST, policy, and NLG modules. Then, we describe the robot's facial expression control and motion control. Finally, we describe and discuss the evaluation results.

\subsection{Ontology of Dialogue Act}
\label{sec:ontology}
An ontology for DA (hereafter, DA ontology or ontology) is used to define DA, which consists of an \emph{intent} and zero or more \emph{slot-value pairs}; an intent represents the intention of the customer or system, and slot-value pairs are detailed information to supplement the intent. We defined the ontology for this dialogue task by referring to \cite{liu2021robustness}. Twenty-three intents and 58 slots were defined. Table \ref{tbl:all_system_intents} shows all the intents defined for the customer and system. For example, {\tt recommend\_target} indicates an action in which the system recommends a certain tourist attraction to the customer. In addition, Table \ref{tbl:examples_of_slot} shows examples of slots. In this table, ``value type'' indicates the type of value each slot can take; ``String'' can be an arbitrary string, and ``Categorical'' can be one of a predefined list. For example, the system DA that tells the customer that there is parking available at a tourist attraction is {\tt inform (attraction\_parking=yes)}.

\begin{table}[t]
\caption{Intents for customer and system defined in DA ontology}
\label{tbl:all_system_intents}
\centering
\begin{tabular}{lp{6.5cm}} \toprule
System & welcome, self\_introduction, finish\_for\_time\_limit, thank-you\_for\_visiting, goodbye, sorry, good, affirm, negate, start\_request, request, confirm\_attraction, inform, recommend\_target, ask\_question, start\_attraction\_introduction \\ \midrule
Customer & thankyou, goodbye, request, affirm, negate, inform, greet \\ \bottomrule
\end{tabular}
\end{table}

\begin{table}[t]
\renewcommand{\arraystretch}{1.3}
\caption{Examples of slot defined in DA ontology}
\label{tbl:examples_of_slot}
\centering
\begin{tabular}{lp{1.8cm}p{3.2cm}} \toprule
Slot & Value type & Description \\ \midrule
user\_name & string & customer's name \\
attraction\_parking & categorical \par (unknown/yes/no) & if parking is available at tourist attraction \\
\bottomrule
\end{tabular}
\end{table}

\subsection{NLU}
\label{sec:nlu}
The task of the NLU component is to predict a customer DA from a customer utterance at each turn. For our NLU, we collected utterance data consisting of customer utterance and customer DA pairs, and we trained the mapping from customer utterances to DAs with a generative model. 

\paragraph{Data collection} First, all possible customer DAs were automatically created from the combination of customer intent and slot-value pairs by using the defined DA ontology (see Section \ref{sec:ontology}). As a result, a total of 1,077 DAs (combinations of intent and slot-value pairs) were created. Then, we had crowdworkers create at least two customer utterances representing each DA. For example, for the dialogue act of {\tt inform (user\_accompany=child)}, the following utterance was collected: {\tt I would like to bring my children to see the sights}.

\paragraph{Training NLU with generative model} We modeled NLU as a sequence-to-sequence problem that generates sequences of DAs from customer utterances using a generative model. Specifically, we fine-tuned the Japanese GPT-2 (336M parameters version)\footnote{\url{https://huggingface.co/rinna/japanese-gpt2-medium}} with the collected data. An input sequence to the model is made by concatenating an utterance and DA with ``[SEP]'' tokens:
\begin{quote}
\footnotesize
{\tt
inform (user accompany=child) [SEP] I would like to bring my children to see the sights.
}
\end{quote}
Note that we tested also with the 1.3B parameters version of GPT-2, but it resulted in lower performance.

\subsection{DST}
The task of the DST is to keep the dialogue state up-to-date. The dialogue state consists of information such as the customer's profile, belief state, the attractions focused on by the customer in the dialogue, and the history of the DAs. We implemented rules for updating the dialogue state on the basis of customer DAs. Specifically, we defined update rules for each customer's intent. For example, when \texttt{inform (user\_accompany=family)} is inputted as a customer DA, the content \texttt{user\_accompany} of the belief state is updated to \texttt{family}.

\subsection{Policy}

\begin{figure}[t]
\centering
\includegraphics[clip, width=0.70\linewidth]{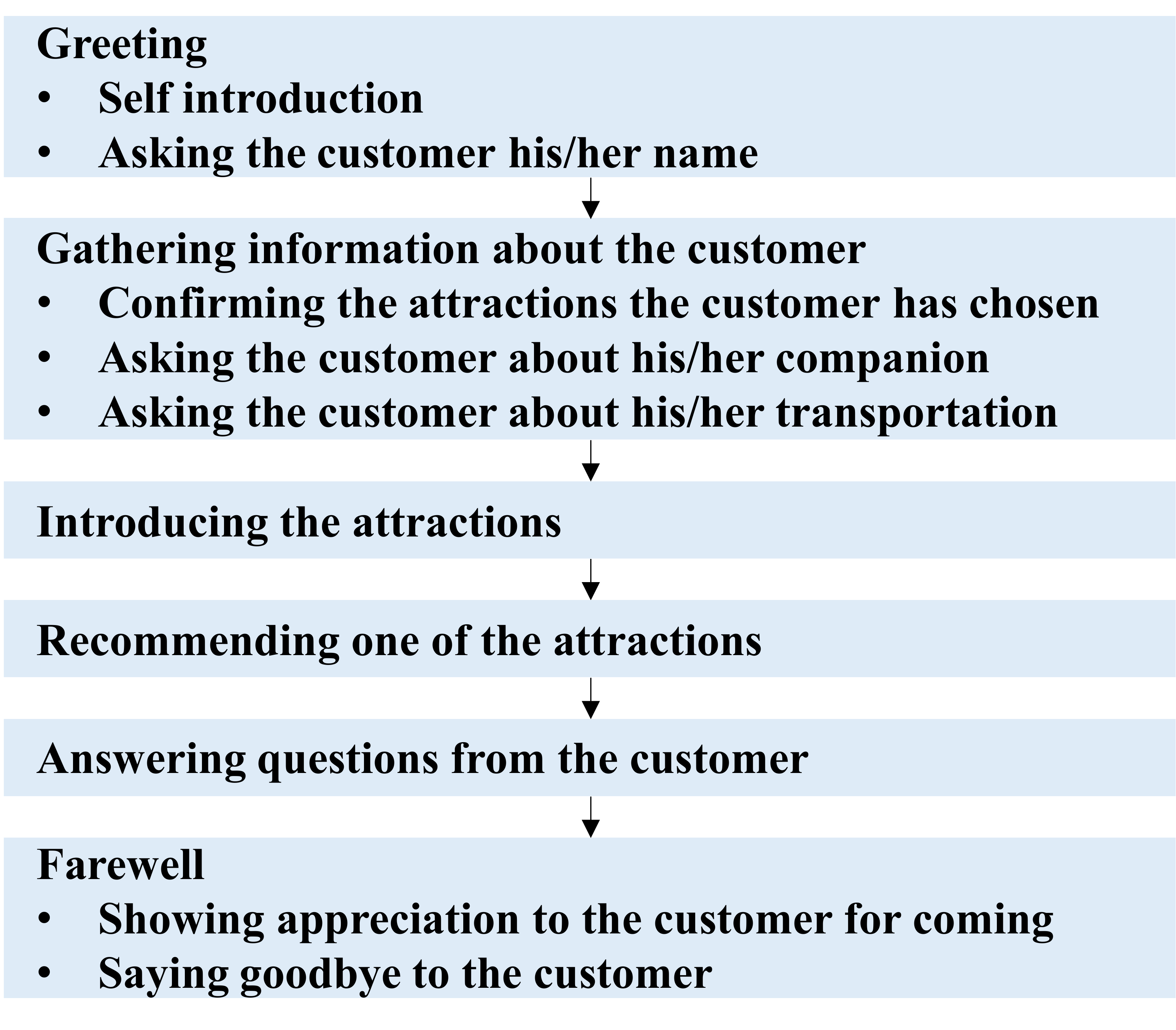}
\caption{Flow of dialogue}
\label{dialogue_flow}
\end{figure}

The task of the policy module is to determine the next system action as a system DA on the basis of the dialogue state updated by the DST. First, we created a dialogue flow that the system should follow (Fig. \ref{dialogue_flow}); the system first greets the customer and asks for information about him/her, such as his/her name, companions, and so on. Next, the system provides a brief description of the two attractions. After that, it presents a recommended attraction. Finally, it answers the customer's questions, says goodbye, and ends the dialogue. On the basis of this dialogue flow, we implemented rules for outputting the system DA on the basis of the current dialogue state so that the dialogue can follow the flow.

\begin{figure}[t]
\centering
\includegraphics[clip, width=0.8\linewidth]{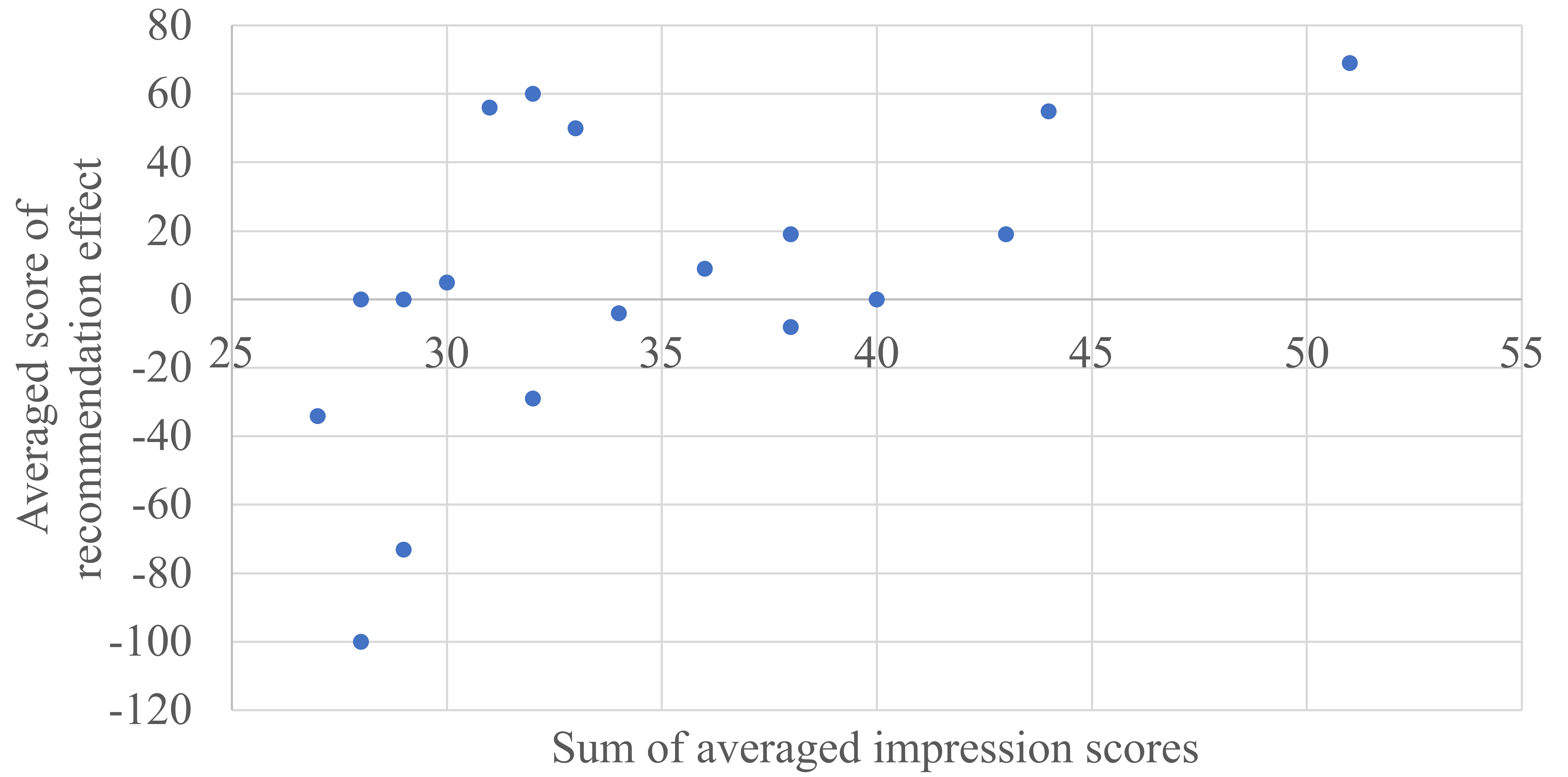}
\caption{Scatter plot of impression scores and recommendation effects}
\label{fig:scatter_plot}
\end{figure}

\begin{table*}[t]
\centering
\caption{Examples of utterances and DAs in dialogue of customer}
\label{tbl:dialogue_1240}
\begin{tabular}{lp{9cm}p{6cm}}
\multicolumn{3}{p{17cm}}{The customer was trying to decide which of two attractions, the Tokyo Trick Art Museum and Tokyo Water Science Museum, to visit. The system was about to recommend the Tokyo Trick Art Museum in this dialogue. The dialogue scored 32 for impression score and -29 for recommendation effect. The utterances were originally in Japanese, but we also show utterances translated into English.} \\
\toprule
Speaker & Utterance & DA estimated by NLU or determined by policy \\
\midrule
System & お客様、どのようなお食事をご希望でしょうか。 \par 
What kind of meal would you like to have? & \texttt{request (user\_food\_type=?)} \\
Customer & 子供が食べやすいものがいいです \par 
I prefer something easy for children to eat. & \texttt{inform (user\_accompany=child, user\_food\_type=steak)}\\
System & ステーキをご希望ですね。かしこまりました。 \par 
You would like a steak, I understand. & \texttt{good (user\_food\_type=steak)}\\
Customer & (silent) &  \\
System & 東京トリックアート迷宮館は雨天でも楽しめるのでおすすめです。 \par 
I recommend the Tokyo Trick Art Museum because you can enjoy it even in the rain. & \texttt{recommend\_target (attraction\_name=Tokyo Trick Art Museum, attraction\_rain=ok)} \\
Customer & (silent) &  \\
System & 何か気になる点やご不明な点はございますでしょうか。 \par 
Do you have any concerns or questions? & \texttt{ask\_question ()} \\
Customer & 何時から何時までやってるんですか \par What are its opening hours? & \texttt{request (attraction\_open\_time=?)} \\ 
System & 東京トリックアート迷宮館は11時から21時まで営業しています。東京都水の科学館の営業時間は9時30分～17時です。 \par 
The Tokyo Trick Art Museum is open from 11:00 am to 9:00 pm. The Tokyo Water Science Museum is open from 9:30 am to 5:00 pm. & \texttt{inform (attraction\_name=Tokyo Trick Art Museum, attraction\_open\_time=11:00-21:00), inform (attraction\_name=Tokyo Water Science Museum, attraction\_open\_time=9:30-17:00)} \\
\bottomrule
\end{tabular}
\end{table*}

\subsection{NLG}
The task of the NLG component is to generate a system utterance from a system DA output by the policy. For our NLG component, the mapping from system DAs to system utterances was trained with a generative model. Data collection and model training were conducted in a manner similar to the NLU implementation (see Section \ref{sec:nlu}). We collected system utterance data for 2,674 automatically created system DAs. For example, for the dialogue act of {\tt request (user\_accompany=?)}, the following utterance was collected: {\tt Who would you like to tour with?} Then, as in the NLU implementation, we fine-tuned the Japanese GPT-2 (1.3B parameters version)\footnote{\url{https://huggingface.co/rinna/japanese-gpt-1b}} with the collected data. 

\subsection{Expression Control and Motion Control}

The robot's expression control and motion control were based on the expression and motion rules defined for each system DA. Specifically, the expression and motion during an utterance and those after an utterance were defined. For example, when the system DA is ``good,'' the robot nods its head with a large smile during an utterance. When the system DA is ``goodbye,'' the robot makes a small smile and bows after the utterance.

\section{RESULTS AND DISCUSSION}
The dialogue performance was evaluated in terms of two factors: the impression evaluation and robot recommendation effect. The impression evaluation consisted of nine questions on a 7-point Likert scale, with a minimum score of 9 and a maximum score of 63 in total. The effectiveness of the robot recommendation was evaluated by the change in the degree to which the customer wanted to visit the sightseeing spot recommended by the robot before and after the dialogue, with a minimum score of \texttt{-}100 and a maximum score of 100. Our system had an impression score of 32 and a recommendation effect of \texttt{-}0.6\footnote{The baseline system developed by the organizers obtained an average score of 38.1 for impression and 5.4 for recommendation effect \cite{drc2022}.}.

Fig. \ref{fig:scatter_plot} shows a scatter plot of impression scores and recommendation effects for the dialogues of our team. The impression score generally ranged from 25 to 45, and the recommendation effects ranged from \texttt{-}100 to 70.

Table \ref{tbl:dialogue_1240} shows a dialogue example of our system. In this dialogue, the system correctly recognized the customer's question, ``What are its opening hours?'' and told the customer the opening hours of the Tokyo Trick Art Museum, which was the most recently mentioned attraction in the dialogue. Furthermore, assuming that the customer would also want to know the opening hours of the Tokyo Water Science Museum, the system also provided the customer with its opening hours at the same time.

However, the customer told the system that she would prefer something easy for the children to eat, but, the system recognized the customer as wanting to eat a steak. This was because the NLU training examples lacked variation and did not contain the phrase ``something easy for children to eat.'' 

Moreover, the dialogue shown in Table \ref{tbl:dialogue_1240} scored a low recommendation effect of \texttt{-}29. The system asked the customer for her food preferences. However, the system could not find restaurants that met the customer's preferences in the vicinity of the attraction and consequently recommended the attraction while ignoring the food preferences she stated. This was because, in the policy module, no rules were defined when the system could not find matching restaurants. Customers' preferences being ignored could have a negative impact on the interest in visiting the recommended attraction.

\section{CONCLUSION}
This paper presented an overview and evaluation results of a dialogue system implemented by the team ``Flow'' in DRC2022. We implemented a pipeline dialogue system consisting of four modules (NLU, DST, policy, and NLG). The results of the preliminary round revealed that our system had limited performance in recommending tourist attractions, and the impression the system had on customers leaves room for improvement. We found that the low variation in training examples for the NLU and failed recommendation due to the policy used were probably the main reasons for the limited performance of the system. In the future, we plan to introduce a method for integrating all of the modules and optimize the dialogue performance of the entire system \cite{ohashi-2022-post}.

\addtolength{\textheight}{-12cm}

\section*{ACKNOWLEDGMENT}
This work was supported by a Grant-in-Aid for Scientific Research (Grant No. JP19H05692).

\bibliographystyle{IEEEtran}
\bibliography{sample}

\end{document}